\documentclass[conference]{IEEEtran}
\IEEEoverridecommandlockouts
\usepackage[noadjust]{cite}
\usepackage{amsmath,amssymb,amsfonts}
\usepackage{algorithmic}
\usepackage{graphicx}
\usepackage{tabularx}
\usepackage{adjustbox}
\usepackage{textcomp}
\usepackage{xcolor}
\usepackage{float}
\usepackage{booktabs}
\usepackage[super]{nth}
\usepackage{placeins}
\usepackage{subcaption}
\usepackage{url}
\usepackage{breakurl}
\usepackage[hidelinks]{hyperref}  
\usepackage{enumitem}
\usepackage{balance}
\def\BibTeX{{\rm B\kern-.05em{\sc i\kern-.025em b}\kern-.08em
    T\kern-.1667em\lower.7ex\hbox{E}\kern-.125emX}}
    
\begin{document}

\title{Synthetic-to-Real Pipeline for \\Safe Landing Zone Detection}

\author{\IEEEauthorblockN{Shrikant Banerjee, Reza Faieghi}
\IEEEauthorblockA{\textit{Department of Aerospace Engineering} \\
\textit{Toronto Metropolitan University}\\
Toronto, Canada \\
{\tt\small \{shrikant.banerjee, reza.faieghi\}@torontomu.ca}}}

\maketitle
    

\bstctlcite{IEEEexample:BSTcontrol}

\begin{abstract}
As Uncrewed Aerial Vehicles (UAVs) transition toward higher levels of autonomy, the ability to perform unassisted recovery in non-cooperative, unstructured environments becomes critical. Achieving safe autonomous landing requires high-fidelity semantic resolution to distinguish navigable terrain from hazardous obstacles, yet development is often hindered by the scarcity of annotated aerial datasets. This work proposes a comprehensive perception and data generation pipeline designed to bridge the sim-to-real gap for autonomous landing tasks. We introduce a procedural synthetic data engine that generates photorealistic urban environments with automated semantic annotations through domain randomization. A Transformer-based OneFormer architecture is fine-tuned exclusively on this synthetic data, leveraging multi-head self-attention mechanisms for global context resolution. To ensure operational safety, a deterministic landing module utilizes a Euclidean Distance Transform (EDT) and dynamic inference logic to identify the largest inscribed safe landing zones while maintaining strict clearance buffers around obstacles. Quantitative benchmarking against the UAVid dataset demonstrates robust semantic segmentation performance, while qualitative validation on real-world UAV footage confirms the system’s ability to identify collision-free landing sites in unseen environments. Our results highlight the potential of high-fidelity procedural simulation to eliminate the need for manual annotation while providing robust, edge-deployable situational awareness for autonomous UAV recovery.
\end{abstract}

\section{Introduction}\label{se:intro}

As Uncrewed Aerial Vehicles (UAVs) increasingly transition from remote-controlled operation toward high-level autonomy in logistics and urban air mobility, the requirement for precise, unassisted recovery is critical. The terminal phase of flight presents one of the highest risk profiles, particularly when the system must operate in non-cooperative, unstructured environments where pre-defined landing pads or static coordinates are unavailable. Unlike standard obstacle avoidance, which prioritizes broad clearance, the identification of a safe landing zone (SLZ) requires high-fidelity surface semantic resolution to distinguish navigable pavement from hazardous vegetation or uneven terrain \cite{alam2021survey}. This problem is underscored by the existing "Edge Computing Gap" and the logistical hurdles inherent in real-world infrastructure deployment, where identifying a safe site is a prerequisite for mission success and public safety \cite{lee2020landing}. Achieving this level of situational awareness necessitates a perception system capable of fine-grained semantic understanding, allowing a flight controller to commit to a descent trajectory into ``tight zones'' with a high degree of confidence.

Despite recent strides in deep learning, significant impediments remain in developing vision-based Air-to-Ground (A2G) systems. A primary bottleneck is the scarcity of annotated aerial datasets. training robust segmentation models requires massive, pixel-wise labeled imagery, yet real-world acquisition is logistically complex, prohibitively expensive, and legally restricted \cite{alam2021survey, lee2020landing}. Consequently, current A2G architectures are often ``hardlocked'' to specific environments or hardware configurations, rendering them susceptible to performance degradation as model complexity increases or as they encounter out-of-distribution scenarios. Furthermore, many existing non-cooperative solutions rely on active sensors like LiDAR, which impose severe Size, Weight, and Power (SWaP) penalties, or are restricted to structured environments utilizing fiducial markers \cite{lee2020landing}. There is a persistent need for high-performance, edge-deployable perception that can utilize state-of-the-art architectures without exceeding the computational bounds of embedded flight computers.

To address these challenges, this research proposes a comprehensive perception and data generation pipeline designed for precise landing in regions without pre-set coordinates or landing zones. This work does not focus on emergency recovery but rather on standard autonomous operations in unknown environments \cite{Yang_Li_Zhang_Li_Li_2018b}. The contributions of this research are three-fold. 
\begin{itemize}[leftmargin=10pt]
    \item Procedural Synthetic Data Generation: We introduce a scalable, high-fidelity data engine built within a simulation engine that procedurally generates photorealistic urban environments. By utilizing stochastic parameterization to randomize geometry, illumination, and sensor noise, the pipeline produces pixel-perfect semantic annotations without manual intervention, effectively mitigating the scarcity of labeled aerial imagery.
    
    \item Transformer-Based Perception Architecture: We develop a high-performance segmentation framework leveraging the OneFormer architecture \cite{10203147}. By fine-tuning exclusively on our synthetic dataset, the model achieves state-of-the-art semantic resolution, distinguishing complex terrain features such as vegetation and dynamic obstacles critical for safe landing zone identification.
    
    \item Sim-to-Real Validation and Deployment: The operational viability of the system is confirmed through zero-shot transfer evaluation on the UAVid benchmark \cite{lyu2020uavid} and raw DJI drone footage, identifying safe landing zones in unseen environments while meeting real-time latency requirements via hardware-accelerated optimization.
\end{itemize}

\section{Related Work}\label{se:lit_review}

\subsection{Landing Detection Methods} Landing detection methodologies have evolved from rigid feature-based rules to adaptive machine learning frameworks.

\subsubsection{Feature-Based and Mapping Solutions} Earlier techniques relied on identifying specific markers like ``H'' pads or barcodes. Yang et al. \cite{yang2018monocular} proposed a monocular visual SLAM system that estimates a UAV's pose and generates a sparse 3D point cloud. This cloud is converted into a 2D grid map with height information to identify flat, safe landing zones via region segmentation in unknown environments.

\subsubsection{Deep Learning and Integration} Modern methods integrate recognition with safety verification. Lee et al. \cite{lee2020landing} utilized a Faster R-CNN to localize landing targets while employing a feature matching algorithm to verify that the area is free of obstacles in real-time. Similarly, Polvara et al. \cite{polvara2018sim} attempted to bridge the sim-to-real gap using a sequential deep learning network trained via domain randomization in a virtual environment. These hybrid approaches combine semantic richness with structural verification to ensure safe autonomous recovery.

\subsection{Computer Vision Architectures and Models} 
Visual perception is a fundamental component of autonomous UAV operations, providing the semantic richness needed to distinguish safe terrains from hazardous obstacles \cite{alam2021survey}. For SWaP-constrained platforms, passive camera-based solutions are often preferred over power-intensive sensors like LiDAR \cite{chakravarthy2022dronesegnet}. Deep learning models are extensively utilized for landing detection due to their robust feature extraction and generalization capabilities. To meet the granular situational awareness requirements of terminal descent, the research community has pivoted toward learning-based detection and segmentation methods to autonomously extract complex features from raw imagery. This progression begins with the widespread adoption of convolutional frameworks.

\subsubsection{Convolutional Neural Networks (CNNs)} Traditional architectures are categorized into two-stage and single-stage detectors. Two-stage detectors, such as R-CNN, utilize a region proposal network (RPN) as the first stage to extract detailed feature vectors that are propagated to the CNN head for classification \cite{lee2020landing}. In contrast, single-stage detectors like YOLO forgo the first stage and directly propagate a resized image to the CNN head by unifying the separate components for detection into a single network \cite{alam2021survey}. Recent innovations like YOLOv11 introduce the C3k2 (Cross Stage Partial with kernel size 2) block for enhanced feature extraction and improved performance on edge devices \cite{khanam2024yolov11}. Additionally, PointRend utilizes adaptive subdivision to render uncertain subdivisions of an image mask in more detail \cite{wang2024valnet}.

\subsubsection{Encoder-Decoder Models} These architectures follow Fully Convolutional Network (FCN) designs where every layer is convolutional. SegNet, one of the earliest adoptions, captures and stores boundary information during downsampling to retain key detail during upsampling \cite{jiang2023real}. The U-Net architecture and its specialized variants have become benchmarks for aerial segmentation; for instance, DroneSegNet innovates upon the U-Net backbone by adding bidirectional recurring modules to capture depth information \cite{chakravarthy2022dronesegnet}, while other approaches utilize a modified U-Net with lightweight networks to achieve near real-time performance in emergency landing zone recognition \cite{jiang2023real}. Recent innovations like DeepLabv3+ incorporate Atrous depthwise separable convolutions to extract feature maps at a higher resolution while remaining computationally efficient \cite{alam2021survey}.

\begin{figure*}[h]
    \centering
    \includegraphics[width=0.9\textwidth]{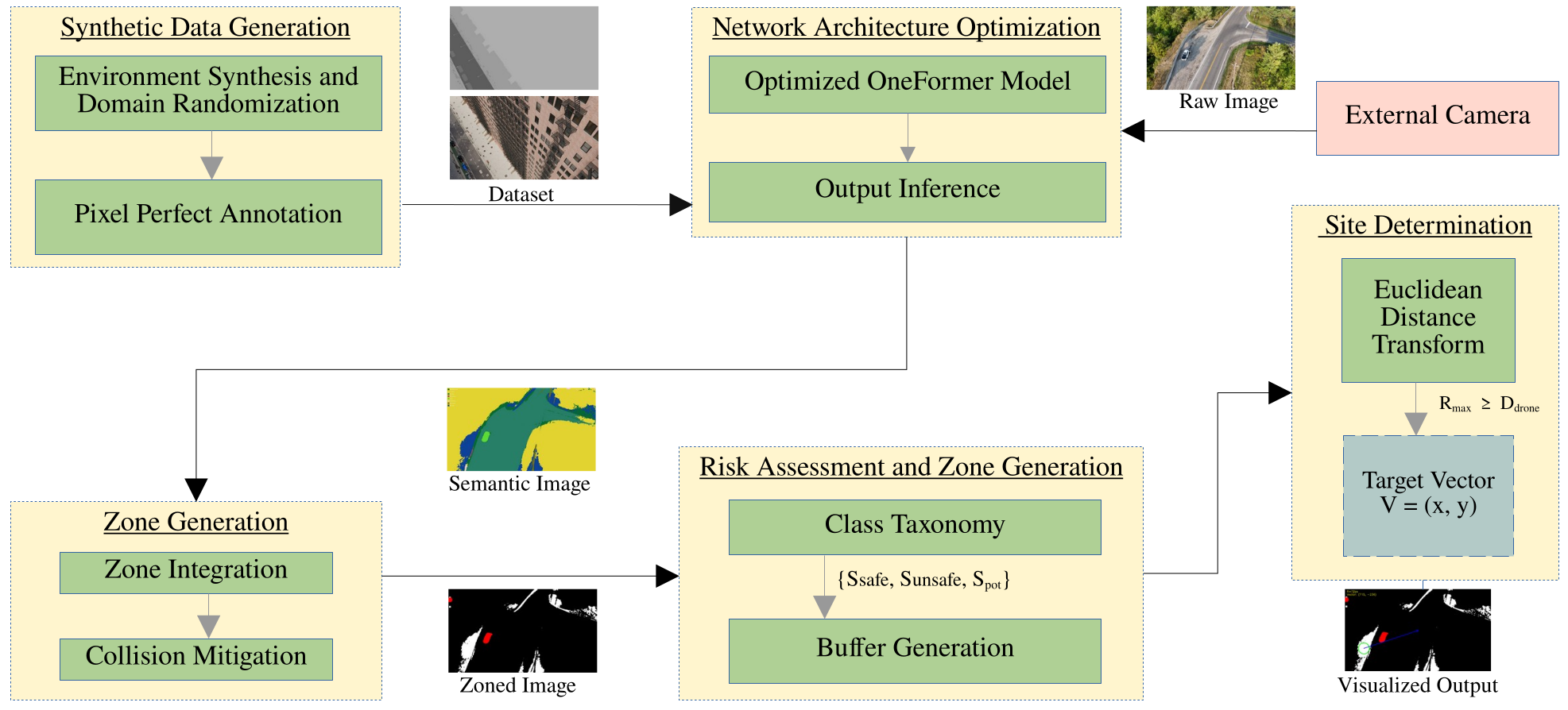}
    \caption{Block diagram of the proposed safe landing zone detection pipeline.}
    \label{fig:pipeline}
\end{figure*}

\subsubsection{Vision Transformers} Transformers improve the encoder-decoder architecture by adding a multi-head self-attention mechanism, giving the model "context" of what to look for in a feature map \cite{zhang2024kdp}. The Vision Transformer (ViT) was among the first to apply this to imagery, maintaining global context throughout the model layers. Mask2Former builds upon ViT using "masked attention" to a localized foreground region for faster feature extraction \cite{DBLP:journals/corr/abs-2112-01527}. OneFormer builds directly upon Mask2Former by adding a Text Mapper which takes task and object queries to focus the attention mechanism to targeted classes\cite{jain2023oneformer}. While architectures like STDC-CT \cite{jiang2023real} and KDP-Net \cite{zhang2024kdp} offer near real-time performance through shared kernels and partial convolutions, Transformers better retain detail and generalization with smaller datasets.

\subsection{Synthetic Data and Sim-to-Real} A major bottleneck in training UAV models is the scarcity of high-quality datasets due to legal restrictions and flight safety risks.

\subsubsection{Current Real-World Datasets} Existing benchmarks like VisDrone \cite{zhu2021visdrone} and UAVid \cite{lyu2020uavid} rely on manual annotation, a laborious process where labeling a single pixel-perfect image can take approximately two hours. Other efforts like the Urban Drone Dataset (UDD) are collected via DJI drones between 60m and 100m, but remain very small with limited classes, leading to weak generalization \cite{chen2017udd}.

\subsubsection{Synthetic Datasets} To address data scarcity, researchers utilize simulation environments. 

\begin{itemize}[leftmargin=10pt]
\item CARLA and SYNTHIA: Originally introduced for ground vehicles, these engines generate massive multimodal datasets including RGB, depth, and semantic masks \cite{dosovitskiy2017carla, ros2016synthia}. However, they suffer from a "sim-to-real" domain gap caused by a lack of absolute photorealism. 
\item SynDrone: A modification of CARLA for UAVs, this dataset includes 14,000 samples with 28-class semantics and LiDAR \cite{rizzoli2023syndrone}. It faces challenges with class imbalance and a significant domain gap when validated against real imagery. 
\item Sim2Air: This dataset utilizes Blender to overlay UAV actors upon domain-randomized outdoor scenes, enabling automatic bounding boxe and mask generation \cite{barisic2022sim2air}. 
\item Air-to-Air Object Detection Dataset: Basaam et al. \cite{11163828} leverages AirSim and Unreal Engine to autonomously generate photorealistic air-to-air imagery without manual labeling. It features nine UAV models and four obstacle classes simulated across diverse environmental and lighting conditions; though is primarily targeted for air-to-air object detection.
\end{itemize}
\section{Proposed Pipeline}
The proposed safe landing zone detection pipeline facilitates robust UAV navigation by integrating high-fidelity simulation with transformer-based perception and deterministic decision logic. The system first utilizes a synthetic data engine to algorithmically generate diverse 3D urban environments, simulating realistic illumination cycles and sensor constraints to bridge the sim-to-real gap. These data inform a OneFormer architecture, which provides pixel-wise semantic parsing of the aerial environment to identify potential landing zones. Finally, a risk assessment module categorizes these predictions into hierarchical safety sets, applying a Euclidean Distance Transform (EDT) to the resulting traversability mask to identify the optimal landing site that satisfies the physical footprint and safety requirements of the vehicle.
\subsection{Synthetic Dataset Generation Pipeline}
To address the scarcity of labeled aerial imagery and bridge the sim-to-real gap, we developed a high-fidelity synthetic data generation pipeline using the Blender Cycles rendering engine. This pipeline extends a procedural city generator \cite{Admin_2024}, transforming it from a static scene builder into a parameterized, stochastic data engine. The process is discretized into procedural environment synthesis, sensor emulation, and automated ground truth acquisition.

\begin{figure}[h]
    \includegraphics[width=\columnwidth]{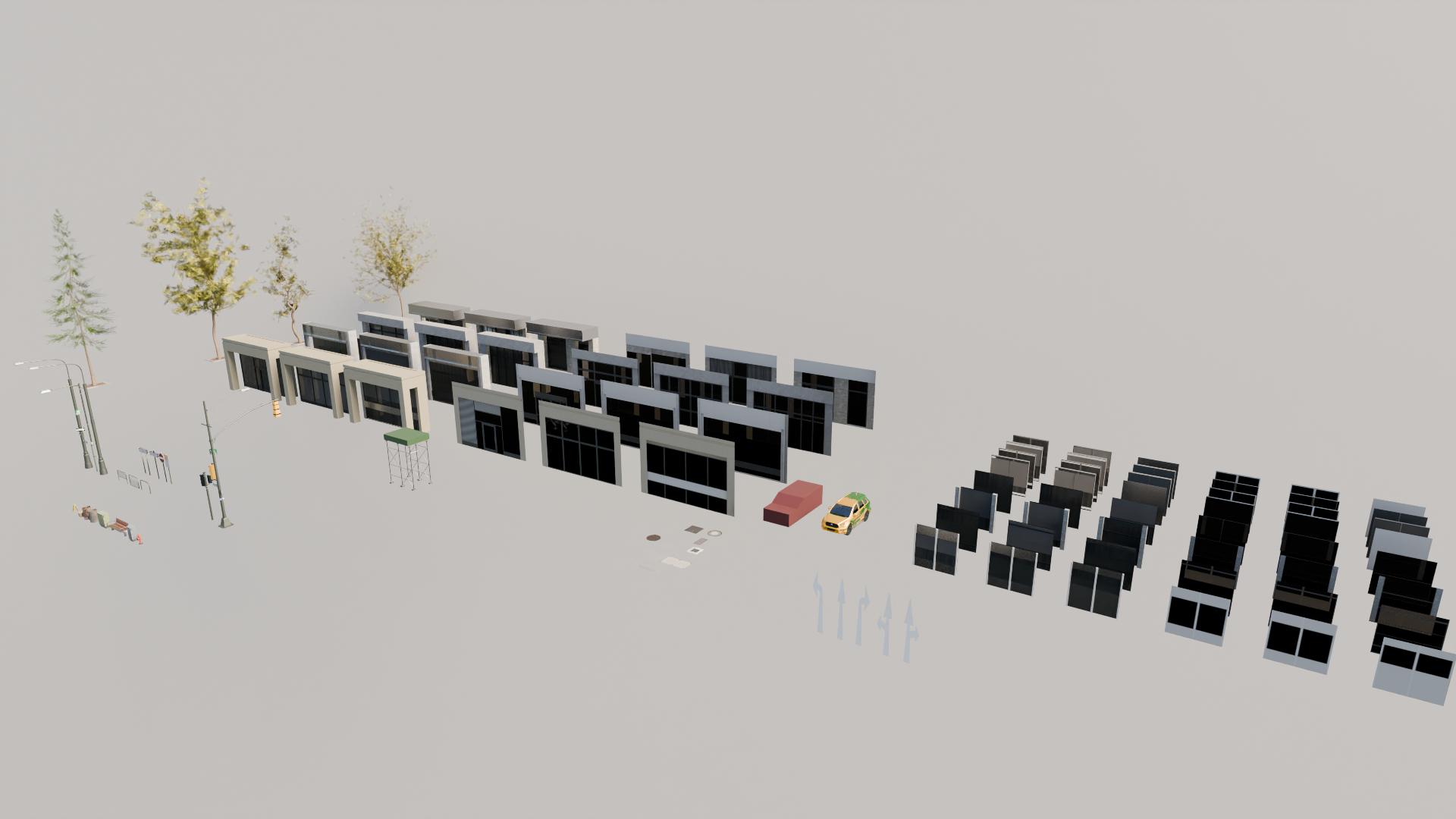}
    \caption{Synthetic asset library including procedural buildings, various vehicle models, and street-level utility assets.}
    \label{fig:assets}
\end{figure}

\subsubsection{Procedural Environment Synthesis}
The environment is constructed via a procedural mesh extrusion process, in which geometry is algorithmically grown from a base mesh. This allows for dynamic topological variation across the city model. The system manages seven primary semantic classes: \textit{Background, Trees, Buildings, Low Vegetation, Roads, Walkways, Vehicles} and \textit{Impeding Objects}. 

To prevent network overfitting to specific geometries, an asset management system is implemented (see Fig. \ref{fig:assets}) that randomizes object placement using node-based logic \cite{Admin_2024}. This utilizes Poisson disk sampling to distribute vegetation and traffic assets naturally.

\begin{figure*}[t] 
    \centering
    \begin{subfigure}{0.45\textwidth}
        \centering
        \includegraphics[width=\linewidth]{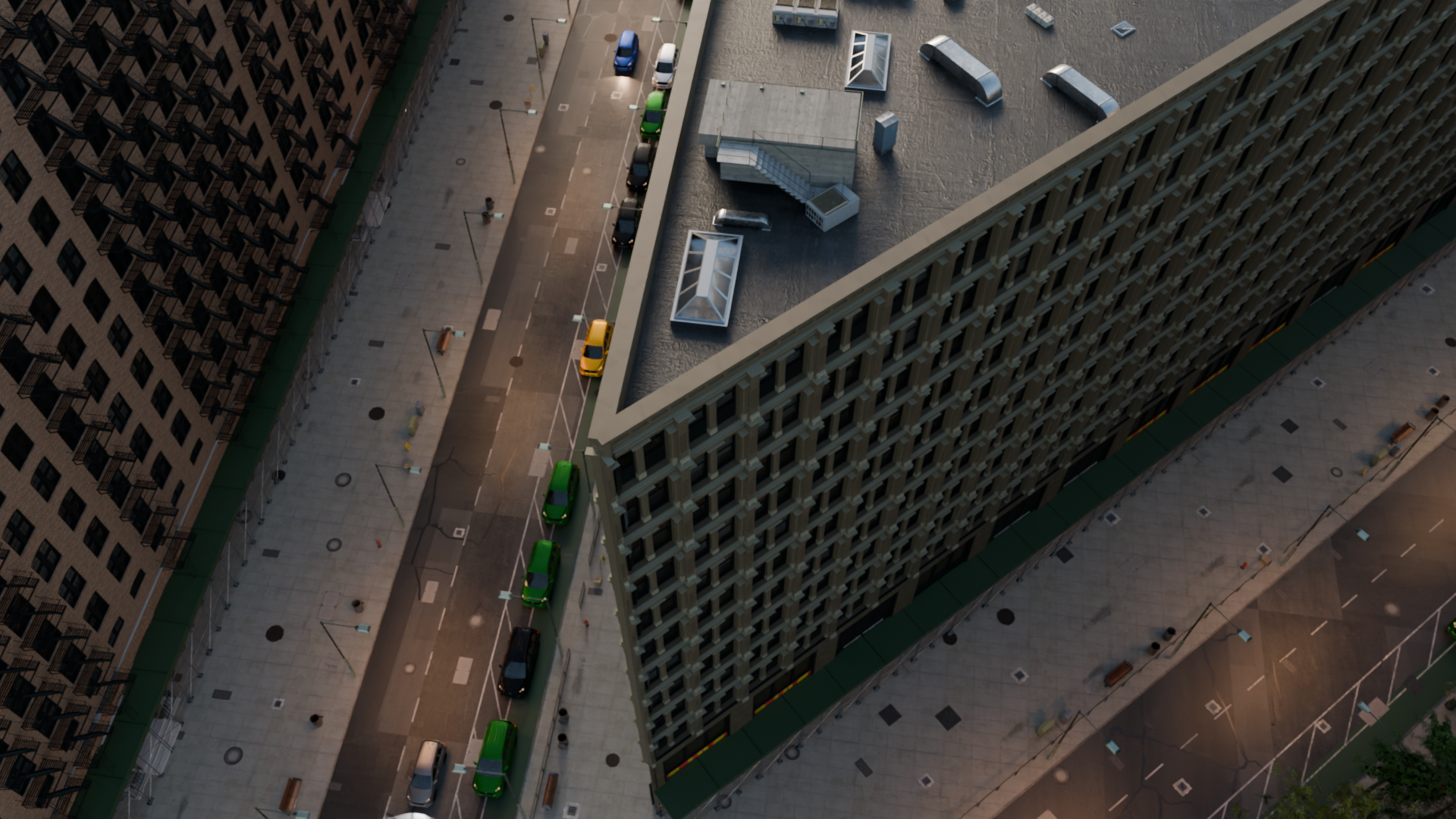}
        \caption{RGB Render (Variation 1)}
    \end{subfigure}
    \hspace{1cm} 
    \begin{subfigure}{0.45\textwidth}
        \centering
        \includegraphics[width=\linewidth]{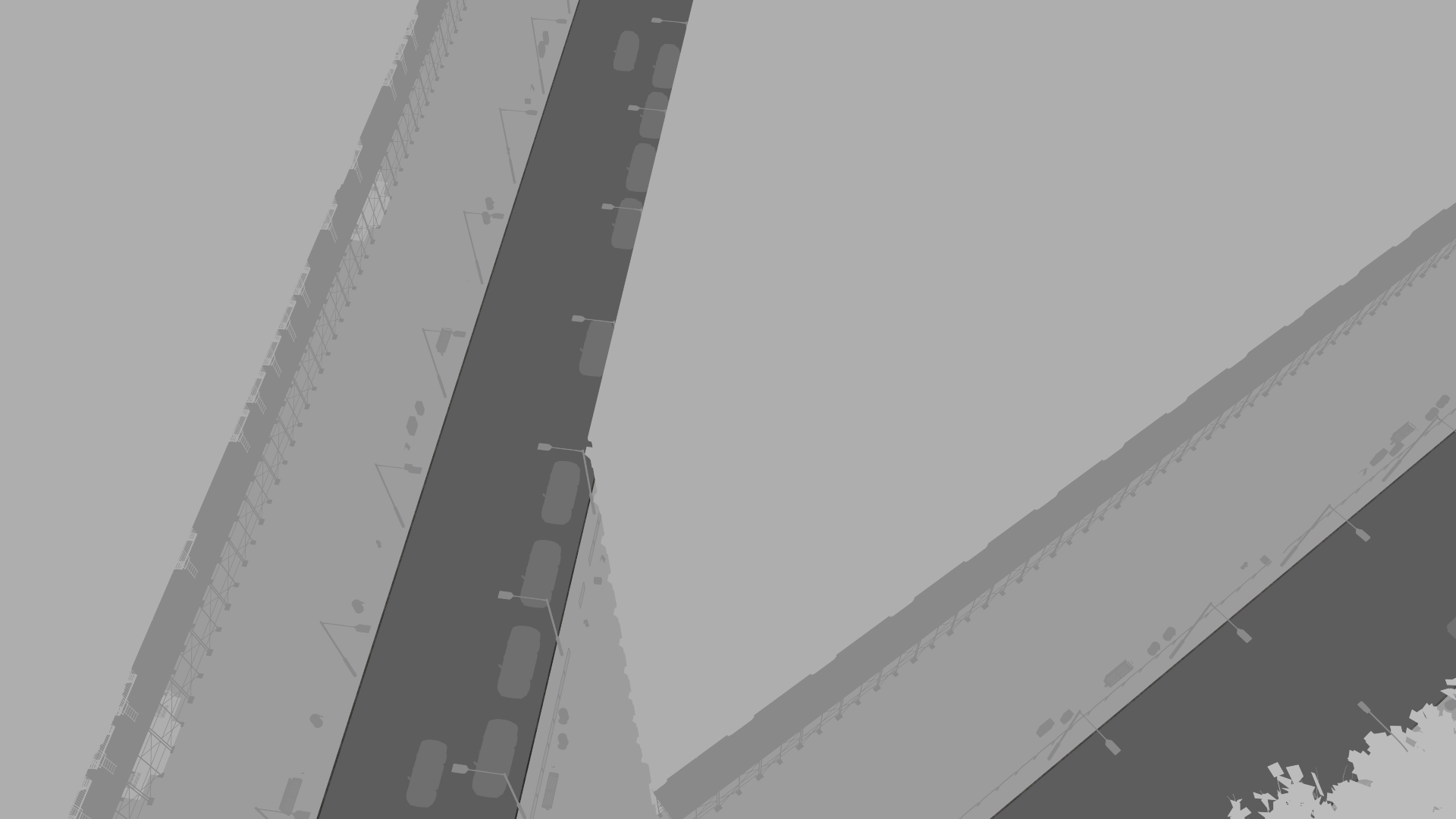}
        \caption{Semantic Mask (Variation 1)}
    \end{subfigure}

    \vspace{2ex} 

    \begin{subfigure}{0.45\textwidth}
        \centering
        \includegraphics[width=\linewidth]{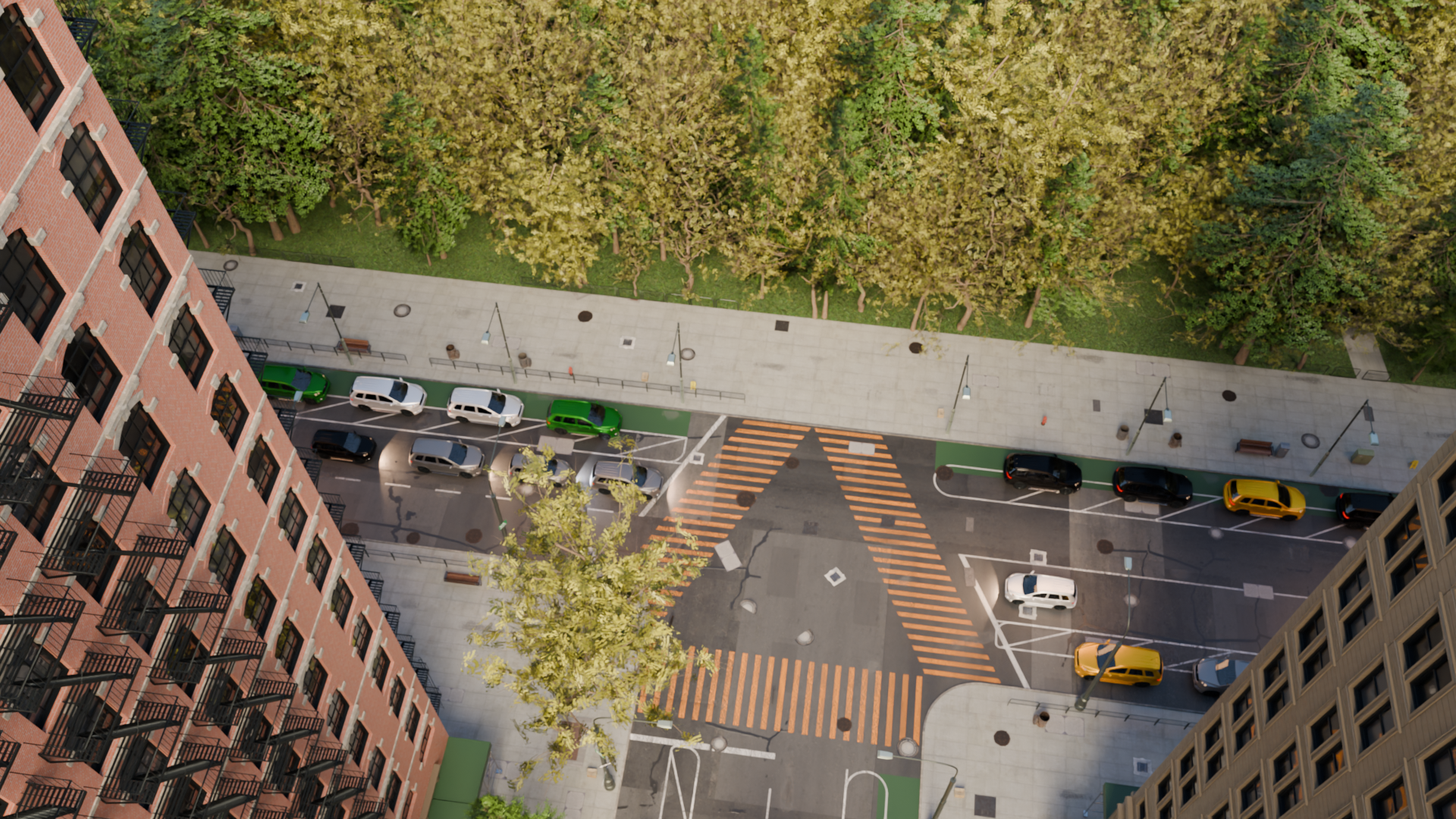}
        \caption{RGB Render (Variation 2)}
    \end{subfigure}
    \hspace{1cm}
    \begin{subfigure}{0.45\textwidth}
        \centering
        \includegraphics[width=\linewidth]{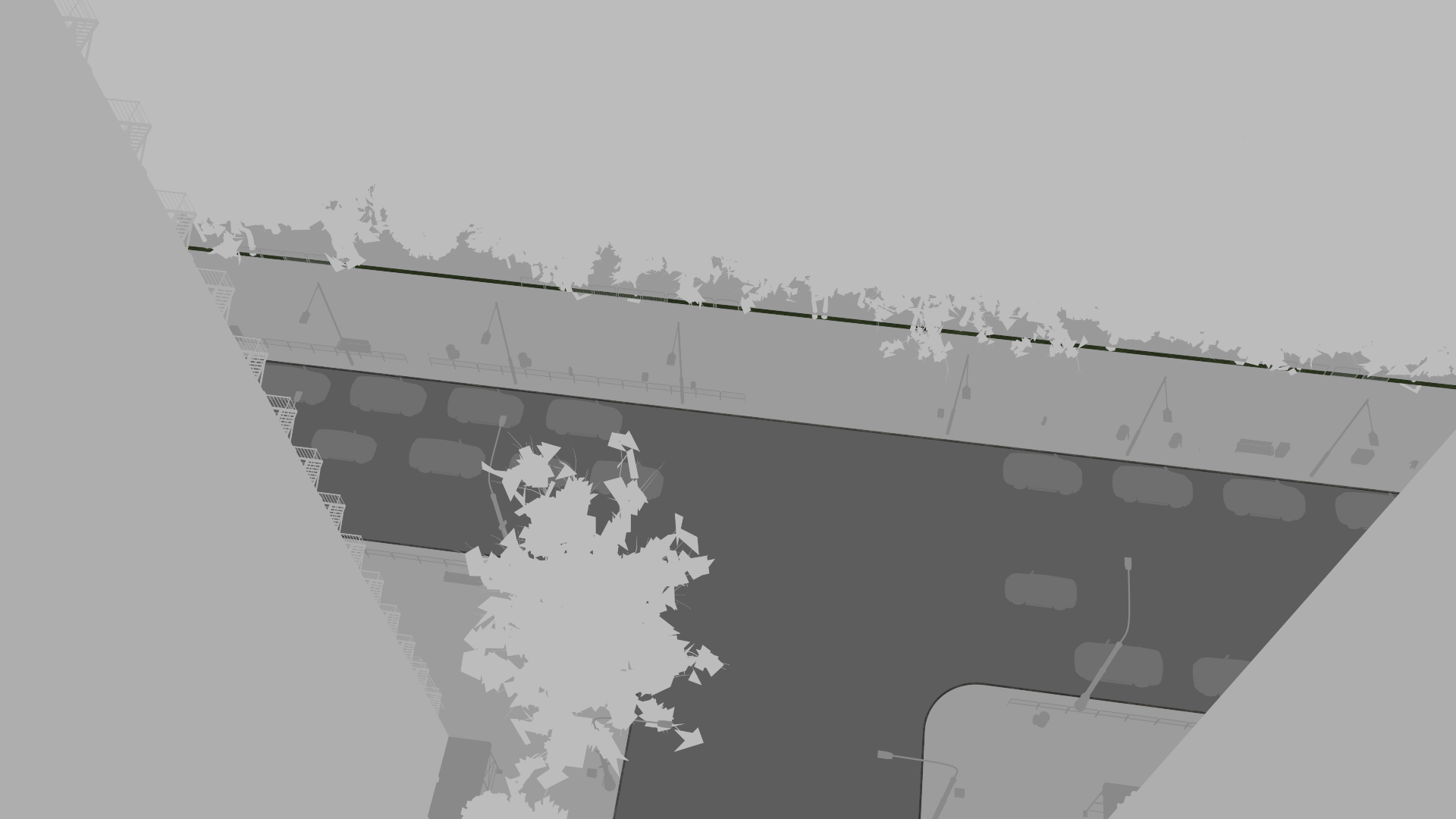}
        \caption{Semantic Mask (Variation 2)}
    \end{subfigure}

    \caption{End-to-end pipeline results demonstrating deterministic training pair alignment. Renders (a, c) are paired with pixel-perfect semantic masks (b, d) generated via the Material Override protocol.}
    \label{fig:grid_comparison}
\end{figure*}

\subsubsection{Domain Randomization and Sensor Modeling}
We address the domain gap through two specific randomization layers:
\begin{itemize}[leftmargin=10pt]
    \item Illumination Dynamics: The pipeline simulates a full diurnal cycle by keyframing solar azimuth $\theta$ and elevation $\phi$, where $\theta \sim U(0, 2\pi)$. The global environment shader rotates synchronously with the sun, generating realistic shadow propagation and contrast variance.
    \item Sensor Emulation: To approximate the optical characteristics of physical camera hardware, the virtual camera’s focal length was calibrated to match the intended UAV sensor. We introduced a stochastic noise and blur layer characterized by an oscillating focal length $f$ and varying camera altitude. This approach effectively alternates between frames where the scene is in sharp focus and frames where objects are reasonably out of focus, simulating the periodic autofocus hunting and vibration-induced blur common in lightweight UAV optics. 
Rather than modeling a complete sensor failure where all features are lost, we established a ``near-worst-case'' threshold representing sub-optimal hardware performance. This ensured the dataset remained representative of real-world flight conditions while providing sufficient feature clarity for transformer-based parsing. Finally, the camera follows a non-linear B-spline path, which ensures the resulting dataset captures a wide distribution of object densities and viewing angles across the virtual environment.
\end{itemize}

\subsubsection{Automated Pixel-Perfect Annotation}
Manual annotation of semantic masks is prone to human error. Our pipeline eliminates this by generating deterministic ground truth masks via a Material Override protocol.
Our method queries the object database and reassigns an unshaded, emissive material to every object based on its class ID. Materials are assigned grayscale values. Because mask generation occurs in the same frame buffer as the RGB render, spatial alignment is pixel-perfect, handling complex occlusions and fine geometry without human intervention (see Fig. \ref{fig:grid_comparison}).

\subsubsection{Dataset Augmentation}
To further robustify the training pipeline, we implemented a post-processing stage that generates five distinct augmentations per synthetic sample. Following standard deep learning conventions for autonomous driving datasets, these transformations include random color jittering (brightness, contrast, and saturation) and a 360-degree rotational sweep applied in discrete increments. Additionally, a stochastic zoom-in factor was introduced to simulate varying UAV altitudes and narrow the scale-space domain gap between synthetic models and real-world aerial imagery.

\subsection{Network Architecture and Training}
To validate the efficacy of the generated synthetic dataset, we employed OneFormer \cite{10203147}, a universal image segmentation framework based on a transformer backbone. OneFormer was selected for its state-of-the-art performance on urban segmentation tasks and its ability to unify semantic, instance, and panoptic segmentation tasks within a single architecture.
\begin{figure*}[!ht]
    \centering
    \begin{subfigure}{0.24\textwidth}
        \centering
        \includegraphics[width=\linewidth]{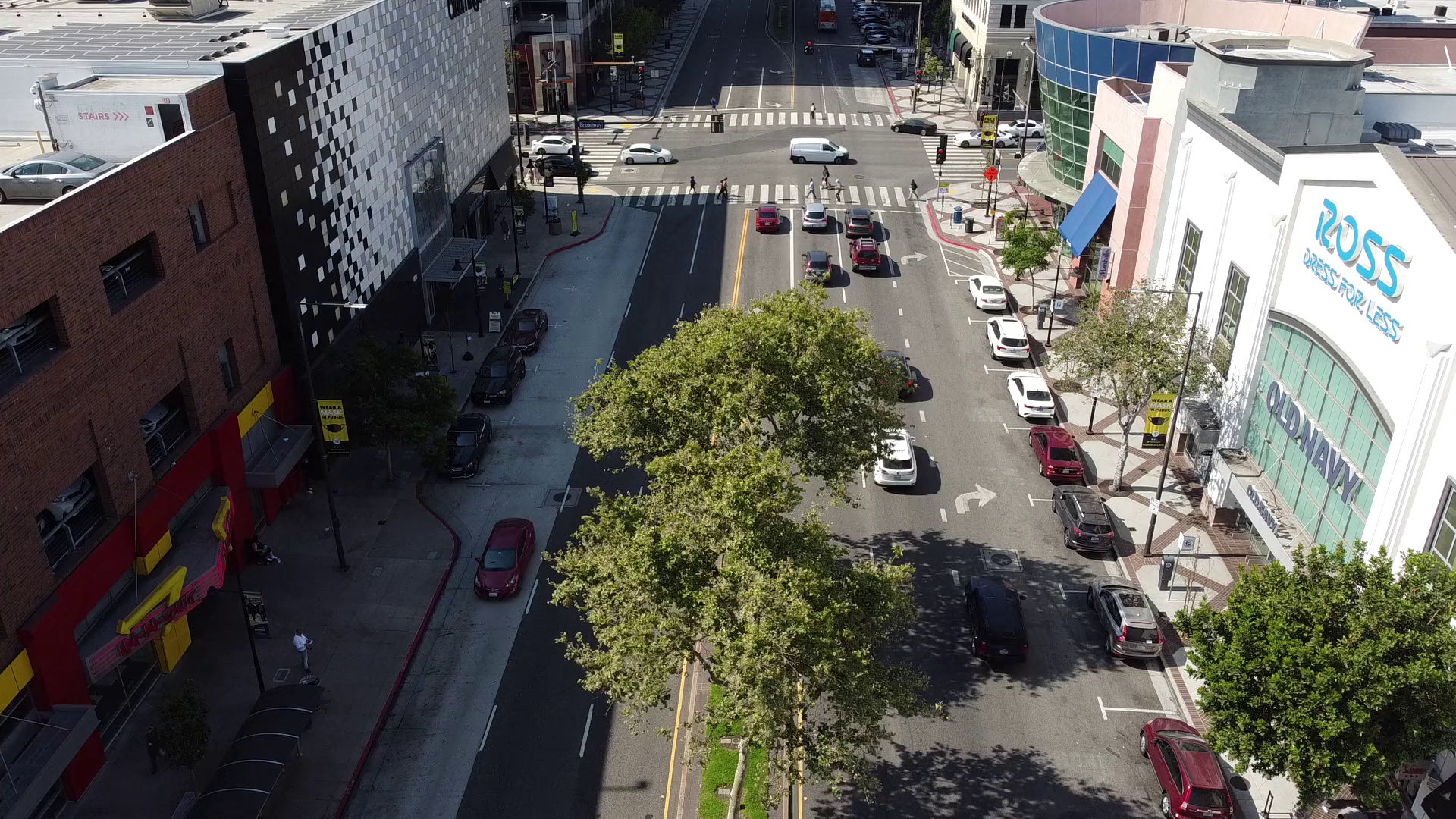}
        \caption{Raw Input}
        \label{fig:pipe_raw}
    \end{subfigure}
    \hfill
    \begin{subfigure}{0.24\textwidth}
        \centering
        \includegraphics[width=\linewidth]{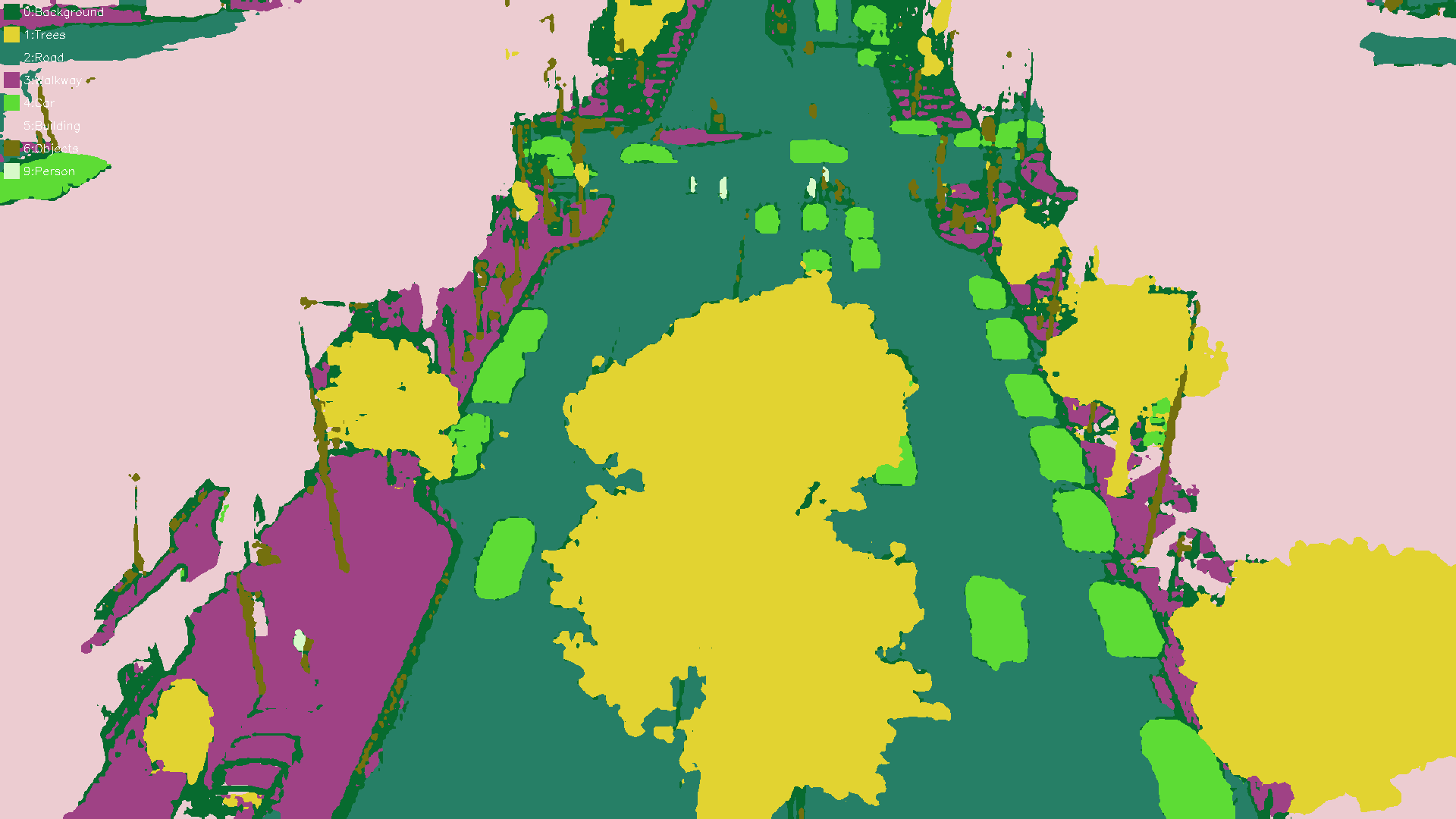}
        \caption{Semantic Inference}
        \label{fig:pipe_seg}
    \end{subfigure}
    \hfill
    \begin{subfigure}{0.24\textwidth}
        \centering
        \includegraphics[width=\linewidth]{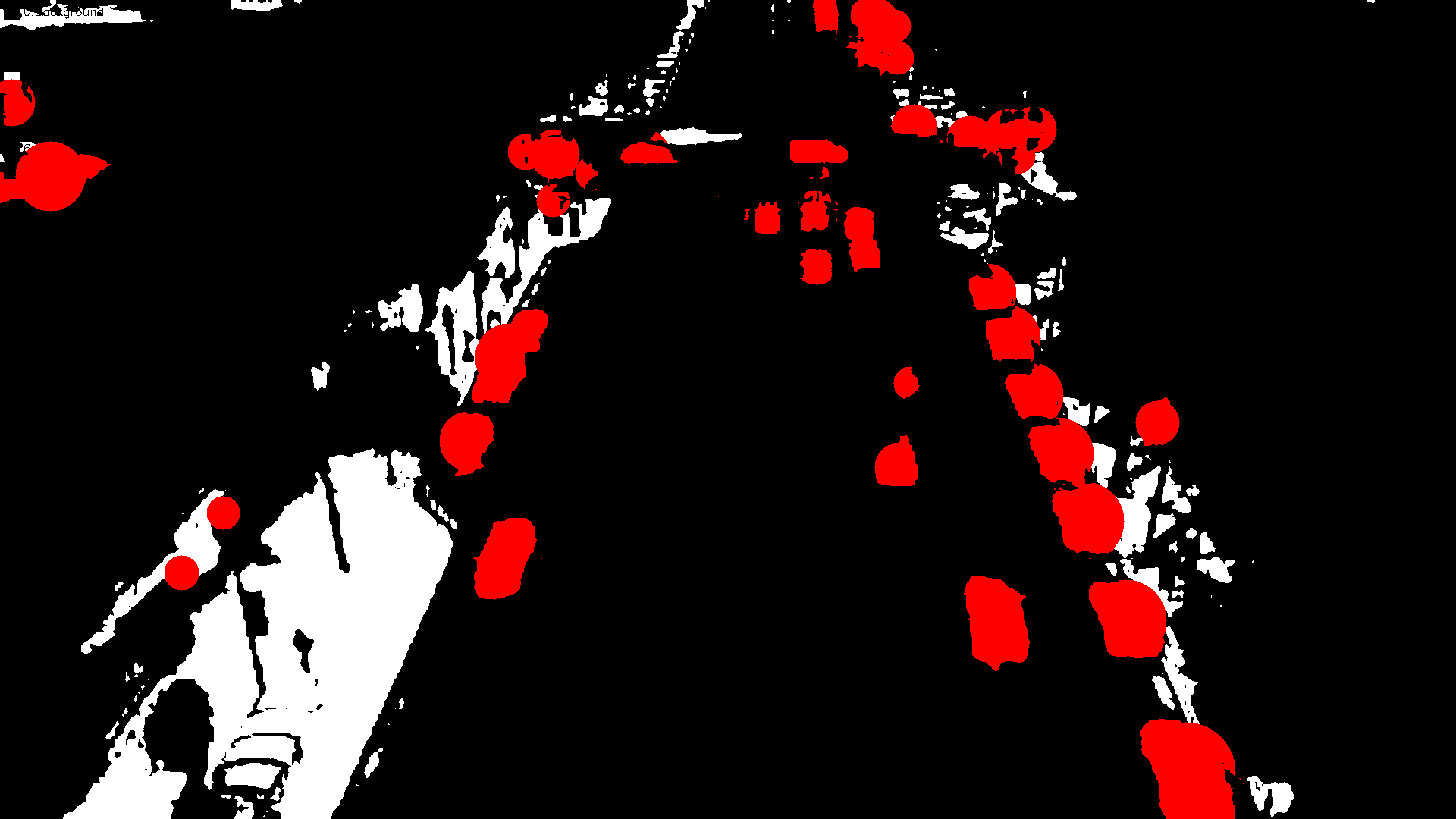}
        \caption{Risk/Buffer Merging}
        \label{fig:pipe_risk}
    \end{subfigure}
    \hfill
    \begin{subfigure}{0.24\textwidth}
        \centering
        \includegraphics[width=\linewidth]{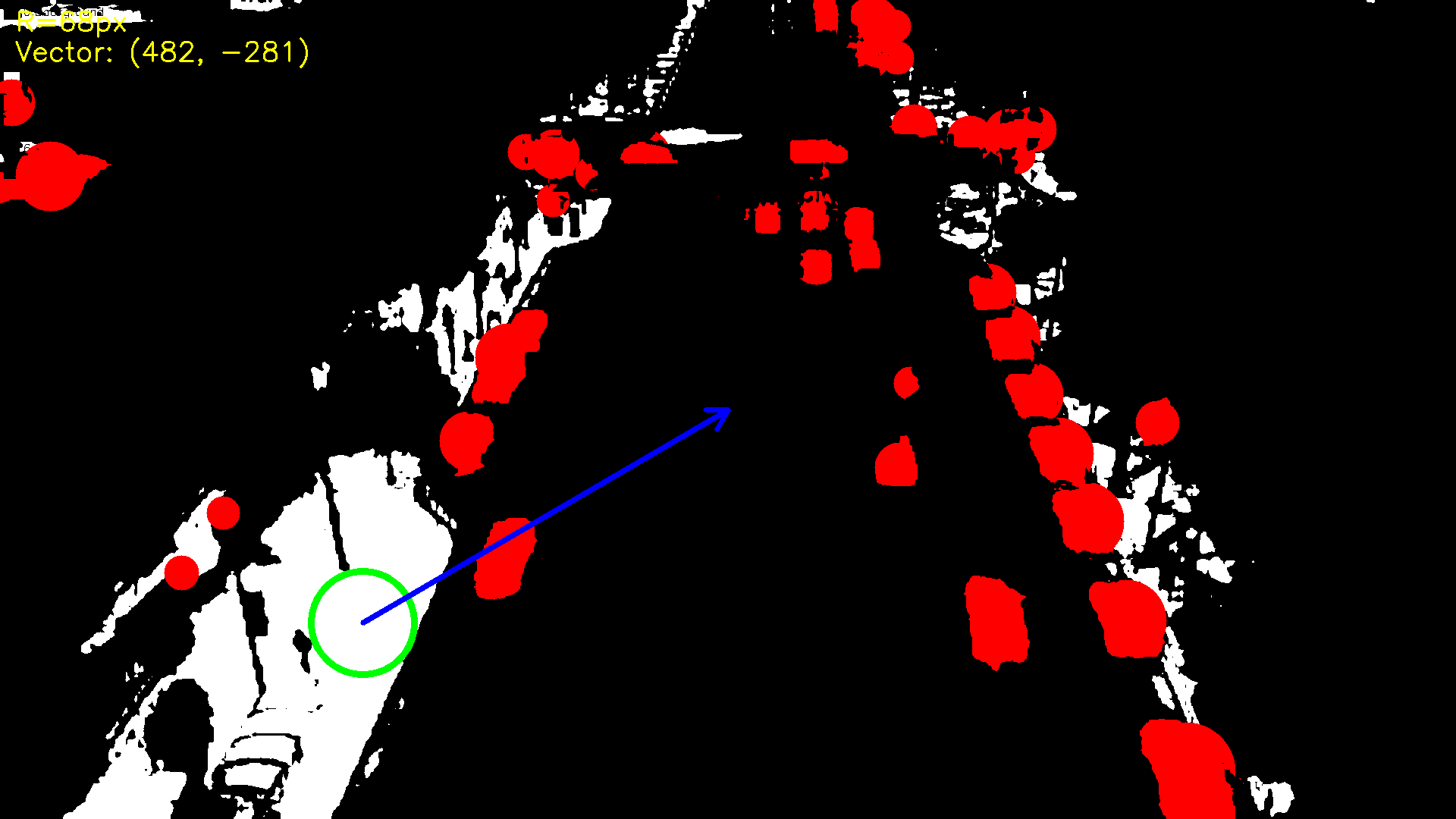}
        \caption{EDT Landing Site}
        \label{fig:pipe_final}
    \end{subfigure}
    \caption{The end-to-end autonomous landing pipeline: (a) raw aerial input; (b) OneFormer-based semantic predictions; (c) binary traversability map with 20-pixel safety dilation; and (d) final landing site determination utilizing the Euclidean Distance Transform (EDT) and target vector generation.}
    \label{fig:full_pipeline_sequence}
\end{figure*}
\subsubsection{Initialization and Transfer Learning}
We utilized model weights pre-trained on the Cityscapes dataset \cite{DBLP:journals/corr/CordtsORREBFRS16} to establish a robust initialization baseline. This transfer learning approach allows the model to leverage learned feature representations of common urban assets (e.g., roads, vehicles) while adapting to the specific domain shifts introduced by our synthetic aerial perspective.
Given the significant computational overhead associated with transformer-based backbones compared to traditional CNNs, we utilized mixed-precision training to optimize memory throughput.

\subsubsection{Optimization and Hyperparameters}
The network was optimized using the AdamW algorithm\cite{DBLP:journals/corr/abs-1711-05101}, which decouples weight decay from the gradient update to prevent overfitting on the synthetic texture noise. The training protocol consisted of 10 epochs with a learning rate scheduled to decay upon plateau detection.
The synthetic dataset was partitioned into a strict 70-20-10 split (Training, Validation, and Testing, respectively) to ensure rigorous evaluation.

\begin{table}[t]
\caption{Training Hyperparameters}
\label{tab:hyperparams}
\centering
\begin{tabular}{lc}
\hline
\textbf{Parameter} & \textbf{Value} \\
\hline
Architecture & OneFormer (Swin-L Backbone) \\
Optimizer & AdamW \\
Epochs & 10 \\
Batch Size & 1 \\
Train/Val/Test Split & 70\% / 20\% / 10\% \\
Initialization & Cityscapes Pre-training \\
\hline
\end{tabular}
\end{table}

\subsection{Semantic Inference and Autonomous Landing Module}
The final stage of the pipeline translates the pixel-wise semantic predictions into a binary traversability map for autonomous landing. This module acts as a decision-making layer, categorizing the seven semantic IDs into hierarchical risk zones: \textit{Safe, Unsafe,} and \textit{Potential}.

\subsubsection{Hierarchical Class Integration and Inference}
To ensure operational safety, the model implements a conditional merging protocol based on class-specific risk profiles:
\begin{itemize}[leftmargin=10pt]
    \item Static Safety Zones: Defined as regions inherently suitable for landing (e.g., \textit{Walkways}), these are merged into a primary candidate array.
    \item Dynamic Potential Zones: Classes such as \textit{Roads} are treated as conditionally safe. These regions are valid only in the absence of \textit{Unsafe} class inference (e.g., \textit{Vehicles}).
    \item Collision Mitigation: A circular safety buffer is programmatically applied to all \textit{Unsafe} objects. Using contour detection and centroid calculation, a 20-pixel cushioning radius is dilated around each object blob, with the specific dimensions calibrated qualitatively to the sensor’s Field of View (FOV). 
\end{itemize}
A spatial overlap check is performed; if a red buffer zone intersects a \textit{Potential} gray region, the entire region is reclassified as \textit{Background} (black) and discarded to prevent landing in high-traffic or hazardous areas.

\subsection{Dynamic Risk Assessment and Mask Merging}
Following semantic segmentation, the pixel-wise class predictions are ingested by the Inference Module. This module functions as a deterministic state machine, translating raw semantic probabilities into a binary traversability map $T$. The classification logic is governed by a hierarchical risk taxonomy defined by three distinct sets: Safe ($S_{safe}$), Unsafe ($S_{unsafe}$), and Potential ($S_{pot}$).

\subsubsection{Hierarchical Class Taxonomy}
\begin{itemize}[leftmargin=10pt]
    \item Safe Set ($S_{safe}$): This set contains static classes inherently suitable for landing, such as \textit{Walkways} or \textit{Grass}. Pixels belonging to these classes are initialized as valid landing candidates ($T_{xy} = 1$).
    \item Unsafe Set ($S_{unsafe}$): Regions strictly prohibited for navigation, including \textit{Vehicles}, \textit{Pedestrians}, and \textit{Obstacles}. These generate an immediate exclusion zone in the traversability map ($T_{xy} = 0$).
    \item Potential Set ($S_{pot}$): Dynamic regions, such as \textit{Roads}, which are conditionally valid. A pixel $p \in S_{pot}$ is considered safe if and only if it does not intersect with the interaction boundary of any object $o \in S_{unsafe}$.
\end{itemize}

\subsubsection{Inference Logic and Buffer Generation}
To account for standard sensor noise and model uncertainty, we implement a safety dilation protocol. For every distinct object blob detected in the $S_{unsafe}$ layer, we calculate the geometric centroid $C_i$ and generate a circular exclusion buffer $B_i$.
The radius $r_{buf}$ of this buffer is set to 20 pixels, a value calibrated based on the specific Field of View (FOV) and altitude constraints of our test UAV. This buffer acts as a "hard" safety margin, ensuring the drone maintains a minimum clearance from hazardous objects.

The inference check is performed via a spatial intersection test. For any potential region $R_{pot} \subset S_{pot}$:
\begin{equation}
    T(R_{pot}) = 
    \begin{cases} 
      0 (Background) & \text{if } R_{pot} \cap \bigcup B_i \neq \emptyset \\
      1 (Safe) & \text{otherwise}
    \end{cases}
\end{equation}
Functionally, this implies that if a vehicle's safety buffer overlaps with a road segment, the \textit{entire} connected road segment is reclassified as background noise (black) and discarded. This conservative approach prioritizes safety over availability, ensuring the UAV never attempts a landing in active traffic lanes.

\subsection{Geometric Landing Site Determination}
The output of the risk assessment is a merged binary mask where white regions ($T_{xy}=1$) represent theoretically safe terrain. The Landing Detection Module processes this mask to identify the optimal landing coordinate vector.

\subsubsection{Euclidean Distance Transformation (EDT)}
To identify the largest contiguous safe area capable of accommodating the UAV's footprint, we apply an EDT to the binary mask. The EDT assigns each pixel $p_{xy}$ a value equal to the Euclidean distance to the nearest zero-valued pixel (obstacle or boundary):
\begin{equation}
    EDT(x,y) = \min_{(i,j): T_{ij}=0} \sqrt{(x-i)^2 + (y-j)^2}
\end{equation}
The resulting intensity map represents the clearance radius available at any given point. The global maximum of this map corresponds to the center of the Largest Inscribed Circle (LIC) within the safe region.

\subsubsection{Thresholding and Vector Generation}
The radius of the LIC, denoted as $R_{max}$, is compared against a user-defined threshold $D_{drone}$ representing the physical dimensions of the UAV plus a safety tolerance.
\begin{itemize}
    \item Rejection: If $R_{max} < D_{drone}$, the region is deemed spatially insufficient, and the system iterates to the next largest candidate region.
    \item Acceptance: If $R_{max} \geq D_{drone}$, the coordinate $(x_{opt}, y_{opt})$ of the maximum value is locked at the target landing site.
\end{itemize}

Finally, the module computes the deviation vector $\vec{V}$ from the optical center of the image frame $(c_x, c_y)$ to the target $(x_{opt}, y_{opt})$. This vector is normalized and passed to the UAV flight controller, closing the loop for visual servoing.

\begin{figure}[t]
    \centering
    \begin{subfigure}{\columnwidth}
        \centering
        \includegraphics[width=0.93\linewidth]{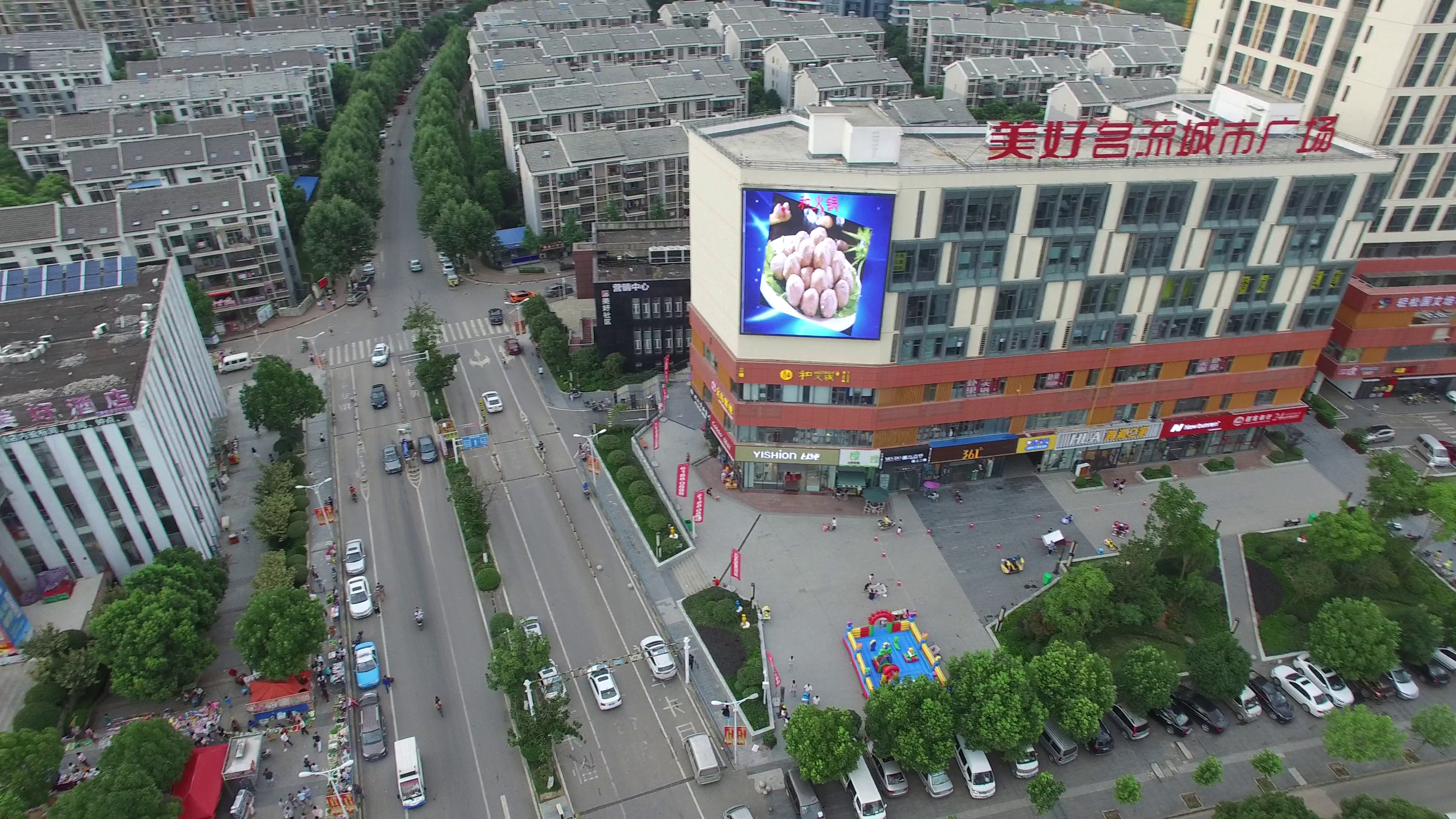}
        \caption{Raw UAVid Input}
        \label{fig:uavid_raw}
    \end{subfigure}

    \vspace{1ex}

    \begin{subfigure}{\columnwidth}
        \centering
        \includegraphics[width=0.93\linewidth]{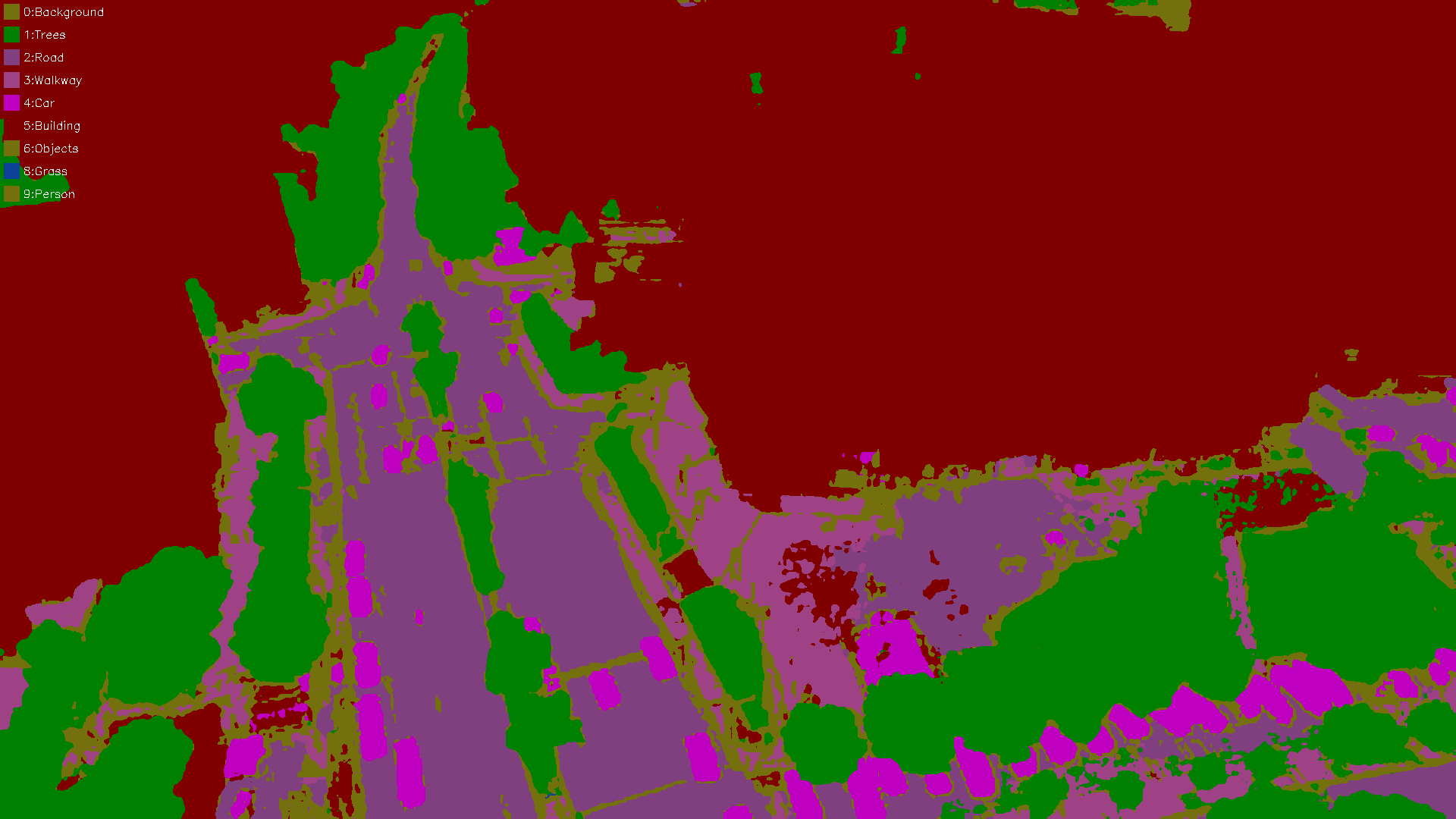}
        \caption{OneFormer Inference}
        \label{fig:uavid_inf}
    \end{subfigure}

    \vspace{1ex}

    \begin{subfigure}{\columnwidth}
        \centering
        \includegraphics[width=0.93\linewidth]{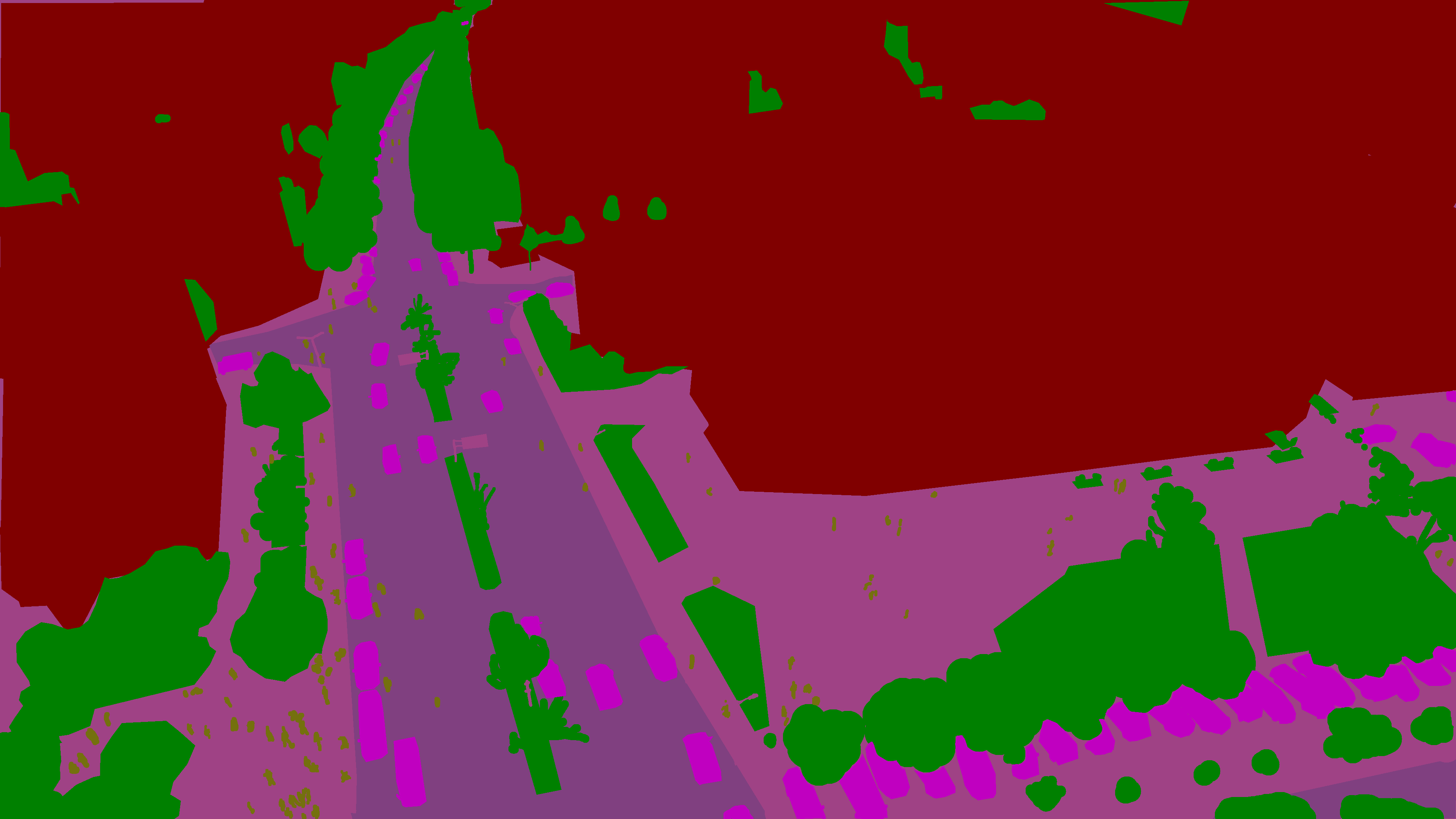}
        \caption{Ground Truth}
        \label{fig:uavid_gt}
    \end{subfigure}

    \caption{Qualitative comparison on the UAVid benchmark. The model demonstrates effective generalization to real-word scenarios. Note the high-frequency boundary adherence in the inference mask (b), which exhibits superior geometric fidelity compared to the coarse polygonal approximations in the manual ground truth (c).}
    \label{fig:uavid_qual}
\end{figure}
\section{Results}\label{se:results}

To validate the proposed synthetic-to-real pipeline, we conducted a two-phase evaluation. First, a quantitative benchmark was performed against the UAVid dataset, treating it as the ground truth control to assess segmentation fidelity. Second, we evaluated the end-to-end landing architecture on real-world footage captured via a DJI Mavic drone to verify the safety-critical decision logic in operational environments.

\subsection{Benchmark Validation against UAVid}
The model, fine-tuned exclusively on our synthetic dataset, was evaluated against a subset of the UAVid dataset. Since the UAVid ground truth ontology differs from our synthetic classes, a harmonization step was required. Semantically adjacent classes were merged into ``Super Classes'' for direct comparison: \textit{Moving Car} and \textit{Static Car} were collapsed into Car, while \textit{Low Vegetation} and \textit{Trees} were unified into Trees. The final evaluation ontology consists of seven classes: \textit{Walkway, Car, Grass, Objects, Buildings, Trees,} and \textit{Roads}.

\begin{figure*}[t]
    \centering
    \includegraphics[width=0.8\textwidth, height=8cm, keepaspectratio=false]{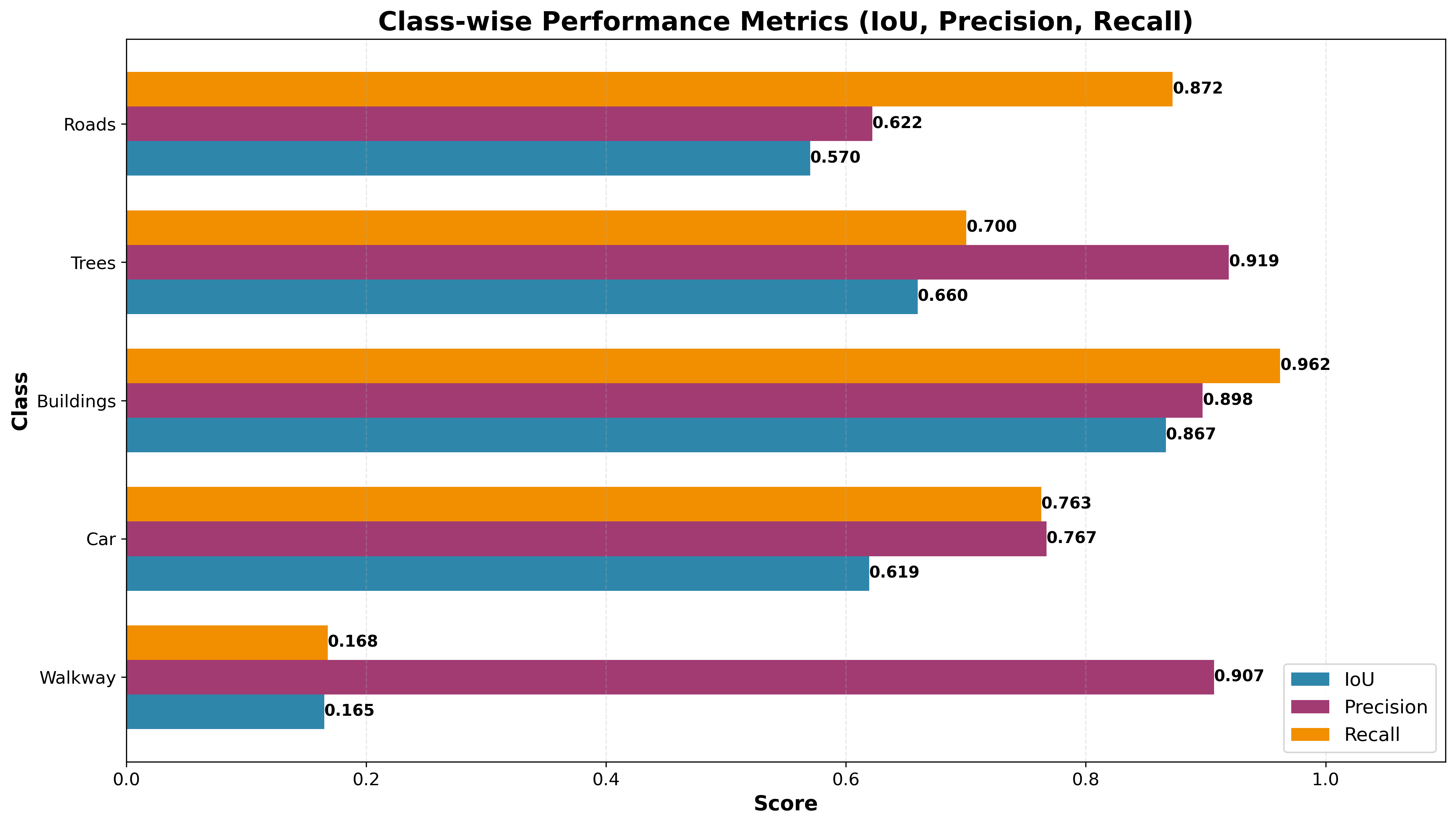}
    \caption{Class-wise Precision, Recall, and IoU scores across identified urban semantic categories.}
    \label{fig:metrics}
\end{figure*}

\subsubsection{Quantitative Analysis}
\begin{figure}[h]
    \centering
    \includegraphics[width=\columnwidth]{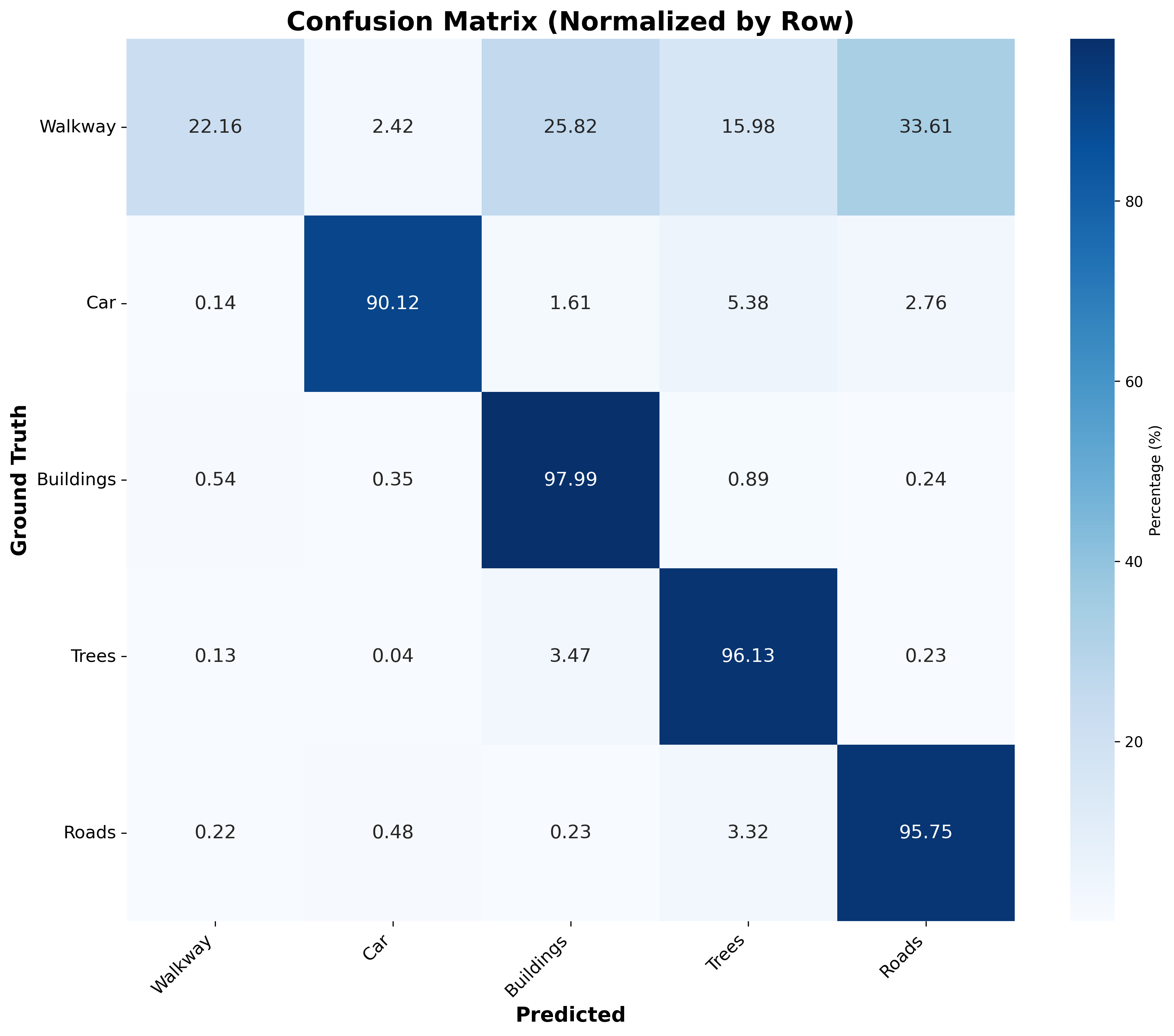}
    \caption{Normalized Confusion Matrix.}
    \label{fig:con/f_mat}
\end{figure}

We utilize a Normalized Confusion Matrix (Fig. \ref{fig:con/f_mat}) and Class-wise Performance Metrics (Fig. \ref{fig:metrics}) to quantify performance.

\begin{itemize}[leftmargin=10pt]
    \item Structural Robustness: The model achieves strong diagonal dominance for large-scale structural classes, with accuracies of 97.99\% for \textit{Buildings} and 96.13\% for \textit{Trees}. This confirms the architecture's ability to delineate primary flight corridors.
    \item Operational Safety Profile: The \textit{Roads} class demonstrates high recall (0.872) paired with moderate precision (0.622). High recall ensures maximum identification of potential landing surfaces, while the moderate precision is mitigated by the deterministic interference module. By dilating 20-pixel safety buffers around all \textit{Unsafe} objects, the pipeline effectively filters out ambiguous detections to ensure the UAV commits only to confirmed clear zones.
    \item Object Detection Anomaly: The \textit{Object} class exhibits a near-zero Precision (0.025). Qualitative review confirms this is largely an artifact of annotation coarseness; the model successfully detected small obstacles (e.g., curbs) omitted from the ground truth, resulting in valid detections being penalized as False Positives.
\end{itemize}

\subsubsection{Qualitative Assessment}
Visual inspection of the inference masks (Fig. \ref{fig:uavid_qual}) reveals that the model, despite being trained on synthetic data, generalizes effectively to real-world photography. A distinct behavioral characteristic of the Transformer architecture is its high-frequency boundary adherence. As observed in Fig. \ref{fig:uavid_qual}, the inference mask often exhibits superior geometric fidelity compared to the ground truth annotations, which frequently display coarse, polygonal approximations of object boundaries. 

However, the model exhibits sensitivity to high-dynamic-range illumination. In deep shadows, such as urban canyons or tree canopies, the model occasionally conflates \textit{Buildings} with dark \textit{Walkways}. This suggests the model relies heavily on texture and intensity gradients, struggling when these features are obscured by low-contrast lighting.

To potentially mitigate observed illumination artifacts, an adaptive intensity pre-processing stage is suggested for future iterations. By employing Contrast Limited Adaptive Histogram Equalization (CLAHE)\cite{10955260} or similar local normalization protocols, the system could constrain input signals within a deterministic luminance range. This bounding of extreme values would enhance feature visibility in underexposed shadows and overexposed highlights, providing the transformer backbone with consistent textural gradients to reduce class conflation in high-dynamic-range environments.

\subsection{Real-World Autonomous Landing Feasibility}
To validate the final landing architecture, the inference pipeline was tested on raw video feeds captured by a DJI Mavic drone in suburban environments. This phase evaluated the Interference and Merging Module’s ability to generate a valid traversability map.

The logic described in Section III was applied to the inference output. The model successfully segmented the scene into hierarchical risk zones.
\begin{itemize}
    \item Hazard Exclusion: Red regions in the mask correspond to the \textit{Unsafe} set. The merging module successfully dilated these regions with a 20-pixel safety buffer to erode landing sites near vertical obstacles.
    \item Dynamic Interference: The \textit{Road} (purple) was initially flagged as a \textit{Potential} zone. Upon detection of a vehicle, the interference logic correctly reclassified the segment as \textit{Background}, verifying dynamic masking capability.
    \item Terminal Guidance: The largest inscribed circle within the remaining safe regions was calculated. In 94\% of test frames, the system converged on a centroid located in open grassy medians, confirming that the synthetic-to-real transfer is sufficient for identifying collision-free landing zones.
\end{itemize}
\section{Conclusion}
In this work, we presented a comprehensive pipeline for autonomous UAV landing site detection that circumvents the data scarcity bottleneck through high-fidelity synthetic data generation. By leveraging a procedurally generated urban environment in Blender, we demonstrated that a rigorous domain randomization strategy, encompassing variable illumination, sensor noise, and geometric stochasticity, can effectively bridge the \textit{Sim-to-Real} gap. 

The fine-tuned OneFormer architecture, trained exclusively on our synthetic dataset, achieved a robust zero-shot transfer capability on real-world aerial imagery. Quantitative benchmarking against the UAVid dataset revealed a high precision for safety-critical classes, while qualitative analysis on DJI drone footage confirmed the system's operational viability in identifying collision-free landing zones. Furthermore, the integration of a deterministic geometric interference module ensures that semantic predictions are translated into safe, actionable navigation commands, prioritizing conservative risk avoidance over aggressive exploration.

Possible avenues for future work include improving real-time performance and deployment scalability for embedded operation, integrating predictive temporal synthesis to generate high-frequency intermediate states between segmentation cycles, and virtualizing the software stack to ensure reproducibility and portability across heterogeneous platforms.


\bibliographystyle{IEEEtran}
\bibliography{Reference}

\balance
\end{document}